\documentclass[runningheads]{llncs}
\usepackage{graphicx}

\usepackage{tikz}
\usepackage{comment}
\usepackage{amsmath,amssymb} 
\usepackage{color}
\usepackage{multirow}

\usepackage[accsupp]{axessibility}  

\begin{document}
\pagestyle{headings}
\mainmatter
\def\ECCVSubNumber{9}  

\title{CNSNet: A Cleanness-Navigated-Shadow Network for Shadow Removal} 


\titlerunning{CNSNet: A Cleanness-Navigated-Shadow Network for Shadow Removal}
%
\author{Qianhao Yu\textsuperscript{$\dagger$}\and
Naishan Zheng\textsuperscript{$\dagger$}\and
Jie Huang\and
Feng Zhao\textsuperscript{$\star$}}
\authorrunning{Q. Yu et al.}
%
\institute{University of Science and Technology of China\\
\email{\{nbyqh,nszheng,hj0117\}@mail.ustc.edu.cn}, \email{fzhao956@ustc.edu.cn}}
\maketitle

\renewcommand{\thefootnote}{\fnsymbol{footnote}} 
\footnotetext[4]{Co-first authors contributed equally.}
\footnotetext[1]{Corresponding author.} 

\begin{abstract}
The key to shadow removal is recovering the contents of the shadow regions with the guidance of the non-shadow regions. Due to the inadequate long-range modeling, the CNN-based approaches cannot thoroughly investigate the information from the non-shadow regions. To solve this problem, we propose a novel cleanness-navigated-shadow network (CNSNet), with a shadow-oriented adaptive normalization (SOAN) module and a shadow-aware aggregation with transformer (SAAT) module based on the shadow mask. Under the guidance of the shadow mask, the SOAN module formulates the statistics from the non-shadow region and adaptively applies them to the shadow region for region-wise restoration. The SAAT module utilizes the shadow mask to precisely guide the restoration of each shadowed pixel by considering the highly relevant pixels from the shadow-free regions for global pixel-wise restoration. Extensive experiments on three benchmark datasets (ISTD, ISTD+, and SRD) show that our method achieves superior de-shadowing performance.

\end{abstract}

\section{Introduction}
“Where there is light, there is shadow.” The shadows, which are prevalent in nature images, often appear when objects partially or completely hinder the light sources. However, undesirable shadows not only fail to satisfy the human perception requirements, but also degrade the performance of the subsequent computer vision tasks \cite{cucchiara2003detecting,jung2009efficient,nadimi2004physical,sanin2010improved,zhang2018improving}, such as object detection, segmentation, and tracking. To improve the human perception and machine perception, it is essential to apply the shadow removal to recover the contents of the shadow regions with the guidance of the shadow-free regions.

As a long-standing computer vision problem, shadow removal has drawn much attention. Existing approaches can be roughly classified into 2 categories: model-based and learning-based techniques. The traditional model-based methods largely depend on the handcrafted priors, e.g., image gradients \cite{finlayson2005removal,gryka2015learning}, illumination \cite{shor2008shadow,xiao2013fast,zhang2015shadow}, and regions \cite{guo2012paired,vicente2017leave}. Due to the limitations of such priors, these algorithms exhibit poor performance when applied to diverse shadow scenes.

Benefiting from the large-scale datasets and the strong learning ability of deep convolutional neural networks (CNNs) \cite{ding2019argan,qu2017deshadownet}, the learning-based methods have provided superior results over the conventional approaches. For instance, Le \emph{et al.} \cite{le2019shadow,le2020shadow} proposed a two-stage network to formulate a linear shadow illumination model to acquire shadow-free images via shadow mattes. DHAN \cite{cun2020towards} applies the dilated convolution to aggregate the multi-context features and attentions hierarchically for artifact-free images. Fu \emph{et al.} \cite{fu2021auto} reformulated the shadow removal as multi-exposure fusion on the multiple estimated overexposed images. However, the convolution operation hinders the CNN-based methods from establishing the long-range pixel dependencies between non-shadow and shadow regions. Hence, these methods cannot fully investigate the information from the shadow-free regions to restore each pixel of the shadow regions. Recently, Chen \emph{et al.} \cite{chen2021canet} explored the potential context relationships between shadow and non-shadow regions, transferring the contextual information from shadow-free patches to shadow patches. Nonetheless, the patch-wise transferring manner is hampered by inaccurate information transformation and complex matching.

Due to the effectiveness of the long-range modeling \cite{vaswani2017attention,wang2018non}, the transformer has recently achieved widespread dominance in many computer vision tasks \cite{chen2021pre,dosovitskiy2020image,liu2021swin,wang2022uformer,zamir2022restormer}. Intuitively, the property of transformer can be utilized to recover the shadow region by establishing the relationship from all the pixels in the non-shadow region to the shadow region. However, the transformer constructs the interaction between all the pixels for recovering the shadow region. Since the features of the shadow region are corrupted while the shadow-free region in the same image has reasonable visibility, only the connection from the pixels with high relevance in the non-shadow region to the shadow region should be formed.

To address the aforementioned issues, we propose a cleanness-navigated-shadow network for shadow removal, namely CNSNet. With the guidance of the shadow mask, the proposed CNSNet investigates the characteristics between the shadow region and the non-shadow region, and establishes the connection from the highly relevant pixels in the non-shadow regions to the shadow region. It comprises three distinct components: a shadow-oriented adaptive normalization (SOAN) module, a shadow-aware aggregation with transformer (SAAT) module, and a soft-region mask predictor. Specifically, for the statistical shift between the shadow-free and shadow regions, the SOAN module performs region-wise restoration by extracting the mean and variance from the shadow-free region and adaptively applying them to the shadow region under the guidance of the shadow mask. This guarantees the statistical consistency between the two regions in a region-wise manner. Furthermore, to build the connection from the non-shadow region to the shadow region, we design the SAAT module with the guidance of the shadow mask. However, the hard shadow mask separates the two regions absolutely, causing the loss of information transmission in the transformer. The soft-region mask predictor is introduced to measure the correlation between the two regions. Therefore, under the guidance of the soft mask, the SAAT module constructs the connection from the pixels with a high correlation in the non-shadow region to the shadow region for pixel-wise restoration.

In summary, our contributions in this work are as follows:

\begin{itemize}
\item We propose a cleanness-navigated-shadow network (CNSNet) for shadow removal, which investigates the relationship from the shadow-free region to the shadow region under the guidance of the shadow mask.
\item We design a SOAN module to extract the statistics (e.g., mean and variance) from the shadow-free region and adaptively apply them to the shadow region for region-wise restoration. Besides, a SAAT module is introduced to take the pixels with high correlation in the non-shadow region into account to recover the shadow region for pixel-wise restoration.
\item Extensive experiments on the public ISTD, ISTD+, and SRD datasets demonstrate that our CNSNet not only achieves competitive results over existing state-of-the-art methods, but also maintains the balance of network parameters, efficiency, and performance.
\end{itemize}

\section{Related Work}
\subsection{Shadow Removal}
Early studies on shadow removal typically make use of various hand-crafted prior information, such as image gradients\cite{finlayson2005removal,gryka2015learning}, illumination properties \cite{shor2008shadow,xiao2013fast,zhang2015shadow}, region characteristics \cite{guo2012paired,vicente2017leave}, and user interactions \cite{gong2016interactive,gryka2015learning,wen2008example}. For example, Finlayson \emph{et al.} \cite{finlayson2009entropy,finlayson2005removal} applied the gradient consistency-based manipulation to recover the shadow images. Shor \emph{et al.} \cite{shor2008shadow} utilized the areas around the shadow edges to estimate the parameters of affine transformations from the shadow to non-shadow regions.

Recently, due to the appearance of large-scale datasets, CNNs have greatly improved the shadow removal performance and gradually become the mainstream of this task \cite{chen2021canet,cun2020towards,fu2021auto,gao2022towards,hu2019direction,le2019shadow,le2020shadow,qu2017deshadownet,wan2022crformer,zhu2022bijective}. For instance, DSC \cite{hu2019direction} creates a direction-aware spatial attention module and aggregates both global and context information. Zhu \emph{et al.} \cite{zhu2022bijective} implemented shadow removal from the perspective of invertible neural networks, and proposed the BMNet with much fewer network parameters and less computational cost. Moreover, generative adversarial networks (GANs) have been widely used in de-shadowing \cite{ding2019argan,hu2019mask,jin2021dc,liu2021shadow,wang2018stacked,vasluianu2021shadow,zhang2020ris}, building on the bidirectional guidance of shadow generation, detection, and removal. ST-CGAN \cite{wang2018stacked} collaboratively detects and removes shadows with the architecture of stacked conditional GANs. G2R \cite{liu2021shadow} employs the shadow generators to synthesize numerous pseudo shadow pairs for joint training. RIS-GAN \cite{zhang2020ris} utilizes the explored relationship among the negative residual images, the inverse illumination maps, and the shadows. Besides, DC-ShadowGAN \cite{jin2021dc} and Mask-ShadowGAN \cite{hu2019mask} exploit adversarial learning and mask-guided cycle consistency constraints and apply unsupervised learning with unpaired datasets.

\subsection{Region-Wise Information}
In recent years, regional information has drawn much attention from researchers in low-level computer vision tasks \cite{chen2021canet,ling2021region,wang2021dynamic,yu2020region,yarlagadda2018reflectance,zhu2020sean}, especially in works related to segmentation or fusion. For example, Ling \emph{et al.} \cite{ling2021region} introduced a region-aware module to develop the visual style from the background and apply it in the foreground, reinterpreting the image harmonization as a style transfer problem. Yu \emph{et al.} \cite{yu2020region} proposed a region normalization, which standardizes the features in different regions during the inpainting network training. DSNet \cite{wang2021dynamic} further combines the deformable convolution with the regional mechanism and dynamically uses region-wise normalization methods for better image inpainting.

In the shadow removal task, previous works have primarily focused on pairing features from the shadow and non-shadow regions. Guo \emph{et al.} \cite{guo2011single,guo2012paired} computed the illumination ratios by randomly sampling pair patches from both sides of the shadow boundary. In \cite{yarlagadda2018reflectance}, shadow detection is treated as a shadow region labeling problem to train a region classifier, and then applies pairs of shadow regions and neighboring shadow-free regions to achieve regional relighting. On the other hand, CANet \cite{chen2021canet} removes shadows by transferring the contextual information of non-shadow regions to shadow regions in a patch-level way.

\subsection{Vision Transformer}
Recently, due to the success of transformer-based models in the field of NLP \cite{vaswani2017attention}, transformer and its variants have widely exhibited outstanding performance in low-level computer vision tasks (e.g., image restoration, enhancement, super-resolution, and dehazing) \cite{chen2021pre,dosovitskiy2020image,liu2021swin,wan2022crformer,wang2022uformer,song2022vision,xu2022snr,zamir2022restormer}. Unlike CNNs, transformer-based network structures are naturally adept at capturing long-range dependencies through the global self-attention. Vision transformer (ViT) \cite{dosovitskiy2020image} is the pioneer in implementing a pure transformer architecture by treating images as token sequences via path-wise linear embedding. For example, IPT \cite{chen2021pre} utilizes typical transformer blocks to train on images with multi-heads and multi-tails for various tasks. Uformer \cite{wang2022uformer} is a hybrid structure consisting of UNet \cite{ronneberger2015u} and transformer for image restoration, with inserted depth-wise convolution in the feed-forward network. Similar to Uformer, Restormer \cite{zamir2022restormer} changes self-attention from spatial dimension to channels, aiming to reduce the computational complexity. Swin transformer \cite{liu2021swin} separates tokens into windows and performs self-attention within a window to maintain the linear computational cost.

\section{Method}
\label{section:method}

\begin{figure}[!t]
\centering
\includegraphics[height=6.5cm]{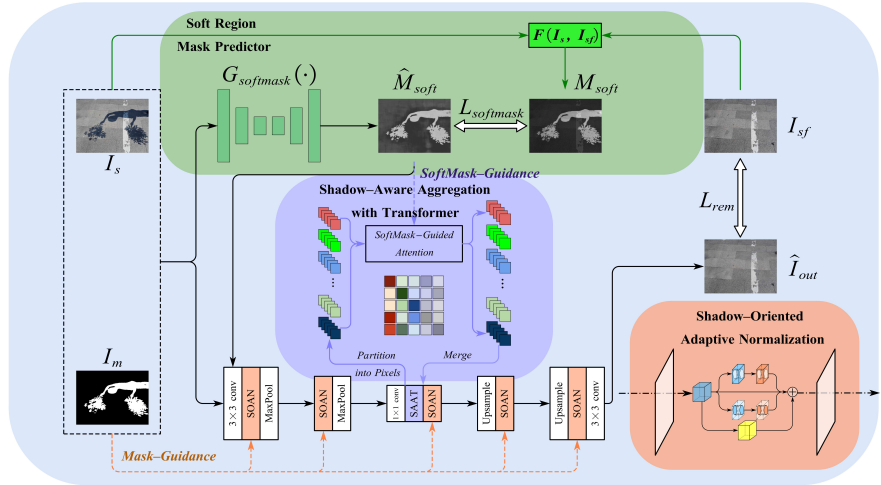}
\setlength{\abovecaptionskip}{-0.01cm}
\setlength{\belowcaptionskip}{-0.2cm}
\caption{Illustration of our proposed cleanness-navigated-shadow network (CNSNet) for shadow removal. It involves three key elements: soft-region mask predictor (green box), shadow-oriented adaptive normalization (SOAN) module (orange box), and shadow-aware aggregation with transformer (SAAT) module (purple box). First, the predictor takes in a shadow image and its corresponding shadow mask to obtain a soft-region mask. Then, both hard and soft masks are concatenated with the input image, entering the UNet-like network to produce the shadow-free results. Note that the guidance (dotted arrows) of both hard and soft masks is applied in the region-wise SOAN and pixel-wise SAAT modules, respectively.}
\label{fig:framework}
\end{figure}

Intuitively, objects from shadow regions and non-shadow regions exist in similar contexts except for illumination, to some extent, making it possible to allow the non-shadow regions to guide the shadow regions. On the basis of this, we elaborate in Sec.~\ref{section:CNSNet} on the overview of our proposed cleanness-navigated-shadow network (CNSNet), which is a composite CNN-transformer framework. With the input shadow image and the corresponding shadow mask, CNSNet consists of three key parts: soft-region mask predictor, shadow-oriented adaptive normalization (SOAN) module, and shadow-aware aggregation with transformer (SAAT) module (see Secs.~\ref{section:softmask}~-~\ref{section:SAAT} for more details).

\subsection{Cleanness-Navigated-Shadow Network}
\label{section:CNSNet}
Traditional networks for image enhancement adopt convolutional operations in the hidden layer. However, most simple convolutional operations focus more on the surrounding pixels and only have a small receptive field to extract local information, which may be inadequate to recover the entire images. Specifically, in the shadow removal task, this local information mainly comes from the regions of the same nature (shadow or non-shadow), while ignoring the association and mutual influence of the shadow and non-shadow regions to a considerable extent.

To address this critical issue, we propose the CNSNet with complementary short-range and long-range communications, fully leveraging the regional information. As illustrated in Fig.~\ref{fig:framework}, our CNSNet is an end-to-end designed framework, including both encoder and decoder procedures during the training process. Besides, the soft-region masks are intermediately produced as supplementary auxiliary information for the network training.

The short-range branch is implemented based on the convolutional and sampling operations during the encoding and decoding procedures, whereas the long-range branch uses a transformer structure to capture the non-local information from the deepest features. In the short-range branch, a novel normalization method SOAN is designed to utilize the non-shadow regional statistics as affine function parameters for the shadow regions after regional instance normalization, thereby roughly ensuring the region-wise statistical consistency. In the long-range branch, the corresponding soft-region mask acquired from the predictor is used to direct the transformer to restore each pixel by taking all the pixels with high relevance into account for the global pixel-wise restoration.

Finally, we employ the pixel-wise $L_1$ distance between our shadow removal outputs $ \hat{I}_{out} $ and the ground-truth shadow-free images $ I_{sf} $ as a loss function $ \mathcal{L} _{rem} $ for shadow removal:
\begin{equation}
\mathcal{L} _{rem}=||\hat{I}_{out}-I_{sf}||_1.
\end{equation}

\subsection{Soft-Region Mask Predictor}
\label{section:softmask}
\begin{figure}[!t]
\centering
\includegraphics[width=8cm]{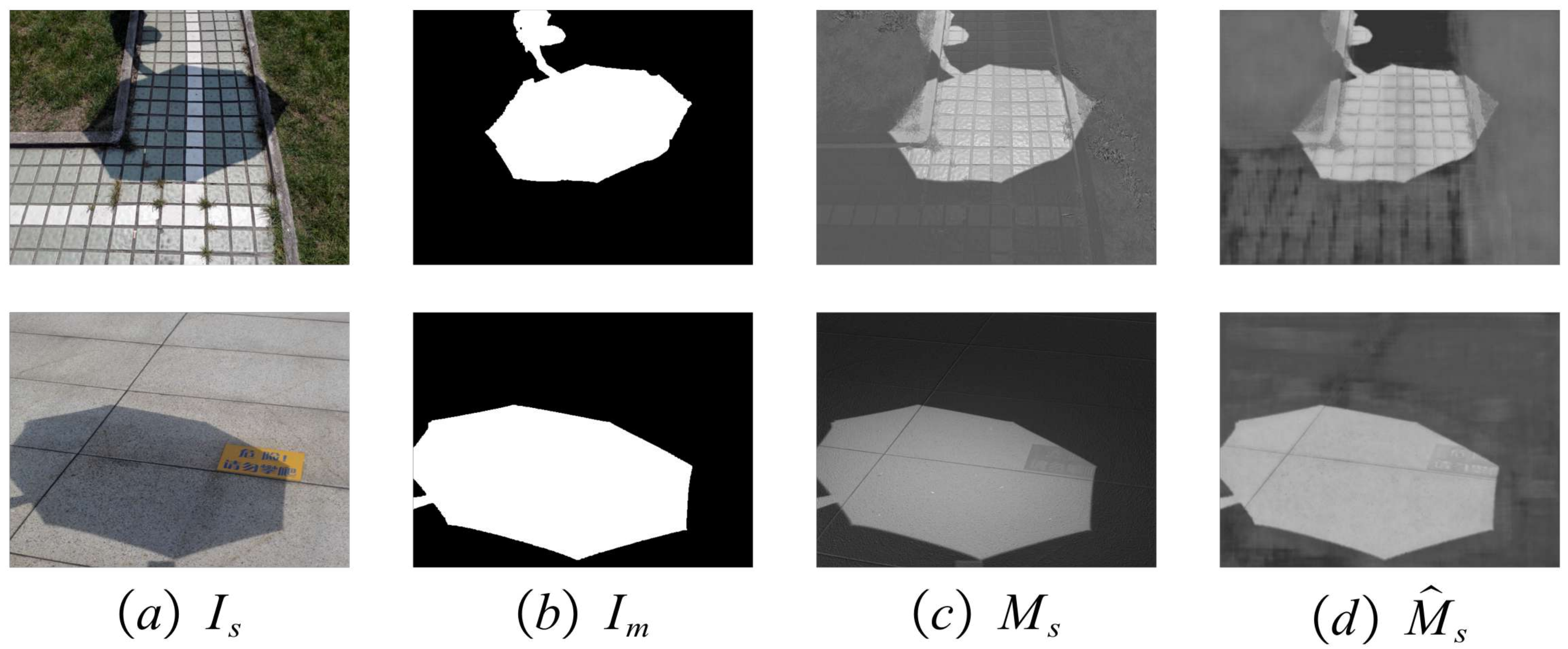}
\setlength{\abovecaptionskip}{-0.01cm}
\setlength{\belowcaptionskip}{-0.2cm}
\caption{Examples of generated soft-region masks. (a) Input shadow image $ I_{s} $, (b) ground-truth hard-shadow mask $ I_{m} $, (c) expected soft-region mask $ M_{soft} $, and (d) generated soft-region mask $\hat{M}_{soft}$.}
\label{fig:eg_softmask}
\end{figure}

Referring to \cite{jin2021dc}, we first compute the difference between the input shadow image $ I_{s} $ and the corresponding shadow-free image $ I_{sf} $ to obtain the expected soft-region mask $ M_{soft} $, and apply the function $ F\left( I_s, I_{sf} \right) $ on the difference:
\begin{equation}
\setlength{\abovedisplayskip}{5pt}
\setlength{\belowdisplayskip}{5pt}
M_{soft}=F\left( I_s, I_{sf} \right) =\frac{1}{3}\underset{c\in \left\{ R,G,B \right\}}{\varSigma}|N\left( I_{s_c}-I_{sf_c} \right) |,
\end{equation}
where $ N\left( \cdot \right) $ is a normalization function on the channel dimension defined as $ N\left( I \right) =\left( I-I_{\min} \right) /\left( I_{\max}-I_{\min} \right) $. Here, $ I_{\min} $ and $ I_{\max} $ are the minimum and maximum values of $ I $, respectively. Note that the values of $ M_{soft} $ are in the range of $ \left[ 0,1 \right] $. Fig.~\ref{fig:eg_softmask} shows some examples of generated soft-region masks.

The network architecture of $ G_{softmask}\left( \cdot \right) $ employs a traditional UNet \cite{ronneberger2015u} structure, combining the shadow image $ I_{s} $ and the hard shadow mask $ I_{m} $ as inputs to generate a soft-region mask $ \hat{M}_{soft} $. As we can see, utilizing $ G_{softmask}\left( \cdot \right) $ can produce a high-quality soft-region mask close to the reference. In other words, the soft-region mask predictor seeks to learn the regional correlation of the shadow and the non-shadow in a fuzzy number-based manner. We explicitly integrate the learned soft-region mask information into the transformer to guide it for better restoration of every shadowed pixel.

During the training phase, we set the predictor to obtain the soft-region mask $ \hat{M}_{soft} $ through the $L_1$ distance loss by:
\begin{equation}
\setlength{\abovedisplayskip}{5pt}
\setlength{\belowdisplayskip}{5pt}
\mathcal{L} _{soft}=||\hat{M}_{soft}-M_{soft}||_1=||G_{softmask}\left( I_s, I_m \right) -F\left( I_s, I_{sf} \right) ||_1.
\end{equation}

\subsection{Shadow-Oriented Adaptive Normalization (SOAN)}
\label{section:SOAN}
Here, we suppose a $ n_{total} $-pixel image with shadows, containing $ n_{shadow} $ shadow pixels and $ n_{non} $ non-shadow pixels. The mean and variance of the two regions are recorded as $ \mu _{shadow} $, $ \mu _{non} $, $ \sigma _{shadow} $, and $ \sigma _{non} $, while $ \mu _{total} $ and $ \sigma _{total} $ represent the statistics of the entire image. Their detailed relationships are as follows:
\begin{equation}
\setlength{\abovedisplayskip}{3pt}
\setlength{\belowdisplayskip}{0pt}
n_{total}=n_{shadow}+n_{non},
\end{equation}
\begin{equation}
\setlength{\abovedisplayskip}{-0.03cm}
\setlength{\belowdisplayskip}{-0.03cm}
\mu _{total}=\frac{n_{shadow}}{n_{total}}\cdot \mu _{shadow}+\frac{n_{non}}{n_{total}}\cdot \mu _{non},
\end{equation}
\begin{equation}
\setlength{\abovedisplayskip}{0pt}
\setlength{\belowdisplayskip}{3pt}
\sigma _{total}^{2}=\frac{n_{shadow}}{n_{total}}\cdot \left( \sigma _{shadow}^{2}+\mu _{shadow}^{2} \right) +\frac{n_{non}}{n_{total}}\cdot \left( \sigma _{non}^{2}+\mu _{non}^{2} \right) -\mu _{total}^{2}.
\end{equation}

Due to the common sense that the values of RGB channels in the shadow region are generally much lower than those in the shadow-free regions, we can observe through the above formulas that both $ \mu _{total} $ and $ \sigma _{total} $ have a large shift compared to $ \mu _{shadow} $, $ \mu _{non} $, $ \sigma _{shadow} $, and $ \sigma _{non} $. Thus, the conventional normalization technique (e.g., BN \cite{ioffe2015batch} or IN \cite{chen2021hinet,ulyanov2016instance}) on the entire image is not competent to overcome this difficulty. In addition, although RN \cite{yu2020region} separately standardizes features based on different regions, where the partition processing is so absolute to ignore any semantic relationship between regions. 

Therefore, we design the shadow-oriented adaptive normalization (SOAN) module. While maintaining the original features, our SOAN utilizes the mean and variance of the non-shadow areas to adaptively assist the recovery of the shadow areas, roughly ensuring the consistency of statistics in the two regions.

\begin{figure} [!t]
\centering
\includegraphics[width=10cm]{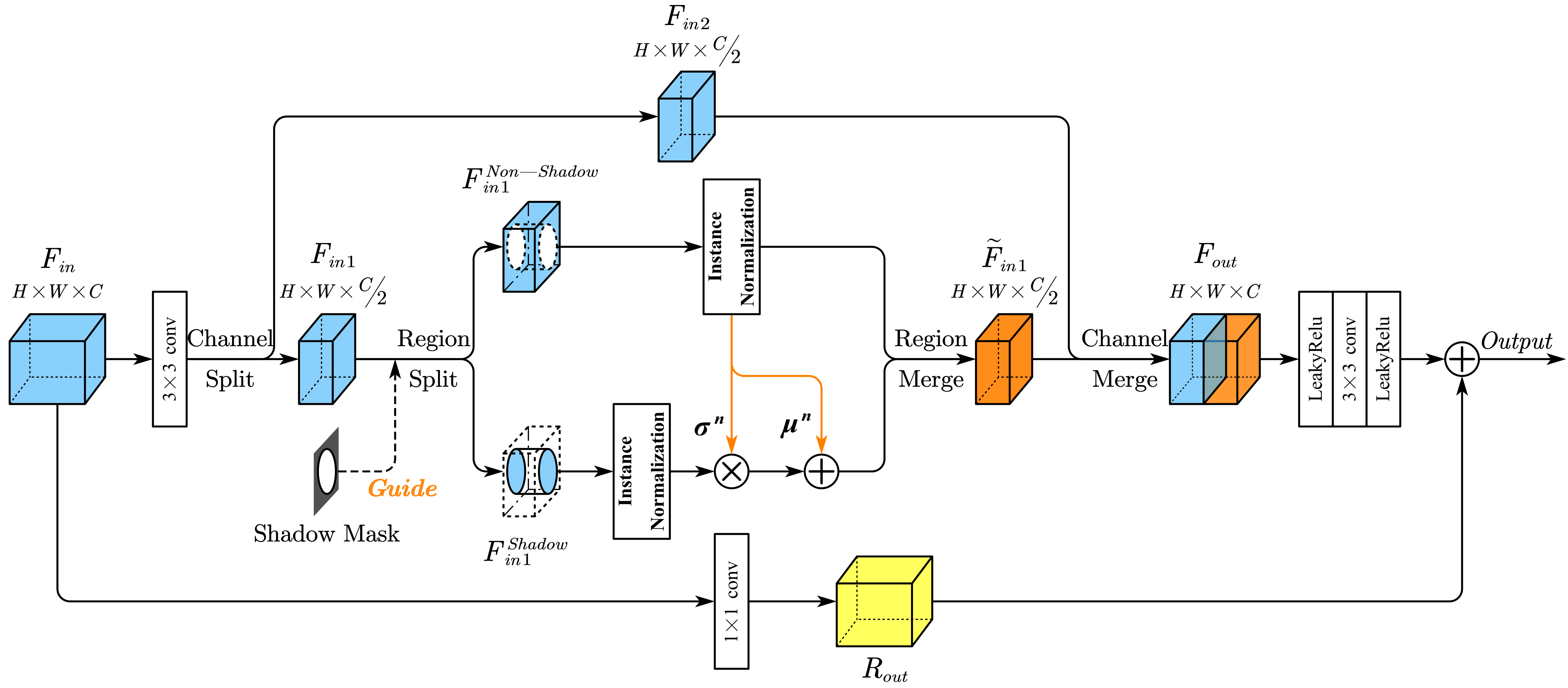}
\setlength{\abovecaptionskip}{-0.01cm}
\setlength{\belowcaptionskip}{-0.3cm}
\caption{Illustration of our proposed shadow-oriented adaptive normalization (SOAN) module. Taking the input features $ F_{in} $ and the corresponding resized shadow mask $ M_{in} $ as the priors, the features are then split across the channel dimensions, and half of them ($ F_{in1} $) performs regional instance normalization, while the other half ($ F_{in2} $) keeps the context information at the same time. Finally, the output is obtained by concatenating the processed features $ F_{out} $ with the residual features $ R_{out} $.}
\label{fig:SOAN}
\end{figure}

As shown in Fig.~\ref{fig:SOAN}, the SOAN block takes the features $ F_{in}\in \mathbb{R} ^{H\times W\times C} $ as inputs and the rescaled shadow masks $ M_{in}\in \mathbb{R} ^{H\times W} $ as prior guidance. $ H $, $ W $, and $ C $ individually denote the height, width, and channels of the current feature maps. Firstly, we divide the convolutional features into two parts on the channel dimension, i.e., $ F_{in1}, F_{in2}\in \mathbb{R} ^{H\times W\times {{C}/{2}}} $. As for $ F_{in1} $, we further split it into two regions: $ R^{shadow} $ (shadow regions) and $ R^{non} $ (non-shadow regions) in a spatial-wise manner according to $ M_{in} $ as below:
\begin{equation}
\setlength{\abovedisplayskip}{5pt}
\setlength{\belowdisplayskip}{5pt}
F_{in1}=F_{in1}^{Shadow}\cup F_{in1}^{Non-Shadow}.
\end{equation}

The two regions are separately standardized by IN \cite{chen2021hinet,ulyanov2016instance} and then re-merge together. Unlike RN \cite{yu2020region}, the IN is used in this case without the learnable affine parameters and the normalized features of shadow regions are affined with the learned scale and bias from non-shadow regions. Specifically, the normalized value of the shadow pixel $ p $ located in $ \left( h,w,c \right) $ can be computed by:
\begin{equation}
\setlength{\abovedisplayskip}{5pt}
\setlength{\belowdisplayskip}{5pt}
\tilde{p}_{h,w,c}=\frac{p_{h,w,c}-\mu _{c}^{s}}{\sigma _{c}^{s}}\cdot \sigma _{c}^{n}+\mu _{c}^{n},
\end{equation} 
where $ p_{h,w,c} $ and $ \tilde{p}_{h,w,c} $ are the initiation and standardization of the pixel value, $ \mu _{c}^{s} $ and $ \sigma _{c}^{s} $ are the channel-wise mean and variance of the shadow features, while $ \mu _{c}^{n} $ and $ \sigma _{c}^{n} $ represent the statistics of the non-shadow regions, calculated by:
\begin{equation}
\setlength{\abovedisplayskip}{3pt}
\setlength{\belowdisplayskip}{-0.03cm}
\mu _{c}^{Region}=\frac{1}{Num^{Region}}\underset{p_{h,w,c}\in R^{Region}}{\varSigma}p_{h,w,c},
\end{equation}
\begin{equation}
\setlength{\abovedisplayskip}{-0.03cm}
\setlength{\belowdisplayskip}{3pt}
\sigma _{c}^{Region}=\sqrt{\frac{1}{Num^{Region}}\underset{p_{h,w,c}\in R^{Region}}{\varSigma}\left( p_{h,w,c}-\mu _{c}^{Region} \right) ^2+\epsilon}.
\end{equation}

Then, $ F_{in2} $ re-concatenates with the normalized $ \tilde{F}_{in1} $ on the channel dimension, which keeps the context information at the meantime. After that, the SOAN module output $ F_{out}\in \mathbb{R} ^{H\times W\times C} $ is integrated through the convolution layers and finally adds with the residual features $ R_{out}\in \mathbb{R} ^{H\times W\times C} $.

In comparison to other normalization methods shown in Table~\ref{table:Ablation_module}, our SOAN is significantly better than single BN \cite{ioffe2015batch} and IN \cite{chen2021hinet,ulyanov2016instance}, further proving the rationality of our aforementioned analysis.

\subsection{Shadow-Aware Aggregation with Transformer (SAAT)}
\label{section:SAAT}
In a variety of image enhancement tasks, traditional transformers \cite{dosovitskiy2020image,vaswani2017attention} can extract non-local information via image patches. However, in general architectures, the attention mechanisms focus on all the patches, which may bring in worthless information. Taking our shadow removal task as an example, to restore a pixel in the shadow regions, the long-range attention may be captured from pixels of both shadow regions and non-shadow regions, while the shadow patches are frequently ineffective. Hence, the inaccurate information brought by the traditional transformer will interfere with the subsequent shadow removal.

\begin{figure}[!t]
\centering
\includegraphics[width=10cm]{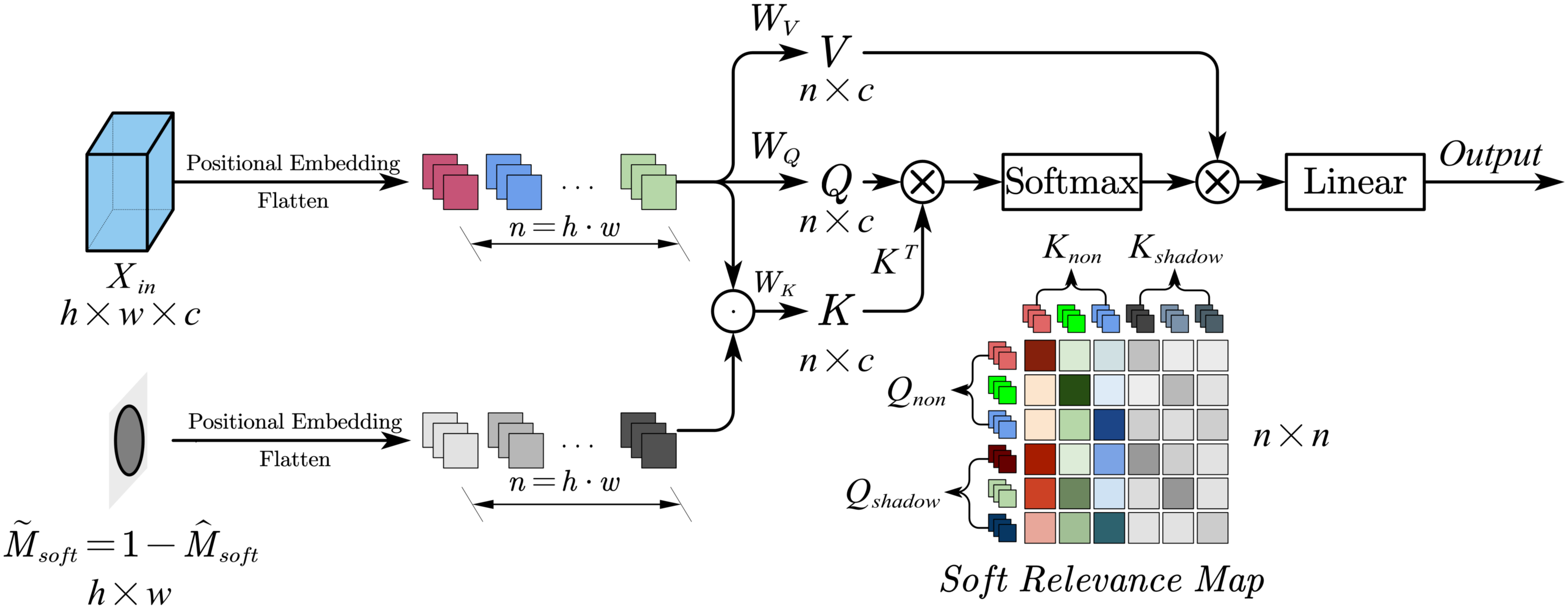}
\setlength{\abovecaptionskip}{-0.01cm}
\setlength{\belowcaptionskip}{-0.3cm}
\caption{Illustration of our shadow-aware aggregation with transformer (SAAT) module with a single head of the transformer layer. The difference from the traditional transformer structure is that we introduce the guidance of a soft-region mask, which multiplies with the input features when calculating the key vectors to acquire the soft relevance map. Based on this map, SAAT produces the outputs in a pixel-wise manner.}
\label{fig:SAAT}
\end{figure}

To this end, we propose a shadow-aware aggregation with transformer (SAAT) module, which improves the performance of this special task by utilizing soft shadow mask-guided attention. The SAAT module consists of two cascading transformer layers, including the multi-head self-attention (MSA) modules and the feed-forward networks (FFNs). Fig.~\ref{fig:SAAT} shows the transformer layer with a single head. Given the input feature maps $ X_{in}\in \mathbb{R} ^{h\times w\times c} $ and associated soft-region mask $ \tilde{M}_{soft}=1-\hat{M}_{soft}\,\,\in \mathbb{R} ^{h\times w} $ acquired from the soft-region mask predictor, we partition $ X_{in} $ and $ \tilde{M}_{soft} $ in a pixel-wise manner, where each pixel is an input token and its channels are token embeddings. Totally, there are $ n=h\times w $ feature patches. Every token executes positional encoding by $ \tilde{x}=x+pe,\ \tilde{m}_{soft}=m_{soft}+pe $, where $ pe $ is the positional embedding. Then, we flatten $ \tilde{X}, \tilde{M}_{soft}\in \mathbb{R} ^{n\times c} $ into 1D vectors and perform the following computation:
\begin{equation}
\setlength{\abovedisplayskip}{3pt}
\setlength{\belowdisplayskip}{-0.03cm}
\tilde{X}=\left[ \tilde{x}_1, \tilde{x}_2,..., \tilde{x}_n \right], \tilde{M}_{soft}=\left[ \tilde{m}_{soft1}, \tilde{m}_{soft2},..., \tilde{m}_{soft\,\,n} \right],
\end{equation}
\begin{equation}
\setlength{\abovedisplayskip}{-0.03cm}
\setlength{\belowdisplayskip}{3pt}
Q=\tilde{X}W_q,\ K=\left( \tilde{X}\cdot \tilde{M}_{soft} \right) W_k,\ V=\tilde{X}W_v,
\end{equation}
where $ W_q, W_k, W_v\in \mathbb{R} ^{c\times d} $ represent the linear learnable matrices, and $ Q, K, V\in \mathbb{R} ^{n\times d} $ are the query, key, and value features, respectively. Here, the soft-region masks affect the values of key features to build better connections between different regions. Following that, we obtain the attention score map $ A\in \mathbb{R} ^{n\times n} $ and the final output features $ \tilde{Y}\in \mathbb{R} ^{\left( h\times w \right) \times d} $ as follows:
\begin{equation}
\setlength{\abovedisplayskip}{5pt}
\setlength{\belowdisplayskip}{3pt}
Attention\left( Q,K,V \right) =softmax\left( \frac{QK^T}{\sqrt{c}} \right) V=AV=\left[ \tilde{y}_1, \tilde{y}_2,..., \tilde{y}_n \right],
\end{equation}
\begin{equation}
\setlength{\abovedisplayskip}{3pt}
\setlength{\belowdisplayskip}{5pt}
\tilde{Y}=FFN\left( \left[ \tilde{y}_1, \tilde{y}_2,..., \tilde{y}_n \right] \right), 
\end{equation}
where $ d $ denotes the number of channels in self-attention computation, which is equal to $ c $ in our design to simplify and keep the same input and output channels. In this case, our design of the transformer layer characterizes the correlation between two regions more accurately with the help of soft-region masks, avoiding the distraction of irrelevant attention. Thus, we ensure that the long-range attentions are from the pixels with sufficient relevance and help to produce global high-quality shadow recovery, as further proved in Table~\ref{table:Ablation_module} of Sec.~\ref{section:Ablation}. 

\subsection{Loss Functions}
Following the previous works \cite{chen2021canet,cun2020towards,fu2021auto,le2019shadow,le2020shadow,liu2021shadow,zhu2022bijective}, except for the $ \mathcal{L} _{rem} $ and $ \mathcal{L} _{soft} $ mentioned above in Sec.~\ref{section:CNSNet} and Sec.~\ref{section:softmask}, we also use a perceptual loss $ \mathcal{L} _{per} $ and a gradient loss $ \mathcal{L} _{grad} $ based on Poison image editing \cite{perez2003poisson}.

Here, $ \mathcal{L} _{per} $ is the perceptual-consistency loss that aims to preserve the image structure with semantic measures and low-level details in multiple contexts. We estimate the feature differences in pre-trained VGG19 networks between the ground-truth shadow-free image $ I_{gt} $ and our shadow-removed image $ \hat{I}_{out} $ as follows:
\begin{equation}
\mathcal{L} _{per}=\underset{k=1}{\overset{5}{\varSigma}}w_k||VGG_k\left( \hat{I}_{out} \right) -VGG_k\left( I_{gt} \right) ||_2,
\end{equation}
where $ VGG_k\left( \cdot \right) $ outputs the multi-scale features of the $ k $-th intermediate layers, and $ w_1 $, $ w_2 $, $ w_3 $, $ w_4 $, and $ w_5 $ are set to 1/32, 1/16, 1/8, 1/4, and 1 in this work.

In addition, $ \mathcal{L} _{grad} $ is proposed by Fu \emph{et al.} in \cite{fu2021auto}, purposing to reduce the gradient domain along the shadow boundary:
\begin{equation}
\mathcal{L} _{grad}=\left( 1-\tilde{M}_{in} \right) \cdot MSE\left( \nabla \hat{I}_{out}, \nabla I_{in} \right) +\tilde{M}_{in}\cdot MSE\left( \nabla \hat{I}_{out}, \nabla I_{gt} \right),
\end{equation}
where $ I_{in} $, $ I_{gt} $, $ \hat{I}_{out} $, and $ \tilde{M}_{in} $ respectively represent the initial shadow images, the ground-truth shadow-free images, our shadow-removed results, and the shadow masks dilated with 7 pixels, and $ \nabla $ denotes the Laplacian gradient operator. It minimizes the gradient domain differences between $ \hat{I}_{out} $ and $ I_{gt} $, while maintaining the gradient domain of non-shadow regions between $ \hat{I}_{out} $ and $ I_{in} $.

In summary, the total loss function of our CNSNet is a weighted sum of the four components described above, which is calculated by:
\begin{equation}
\mathcal{L} _{total}=\lambda _1\mathcal{L} _{rem}+\lambda _2\mathcal{L} _{soft}+\lambda _3\mathcal{L} _{per}+\lambda _4\mathcal{L} _{grad},
\end{equation}
where $ \lambda _1 $, $ \lambda _2 $, $ \lambda _3 $, and $ \lambda _4 $ are hyperparameters to balance different loss terms and are respectively set to 10.0, 5.0, 1.0, and 1.0 in our experiments.

\section{Experiments}

\subsection{Datasets and Evaluation Measurements}
\subsubsection{Benchmark Datasets.}
We utilize three representative public datasets: ISTD, adjusted ISTD (ISTD+), and SRD, to train and evaluate our proposed model. Both the ISTD and ISTD+ datasets contain 1870 image triplets of shadow images, shadow-free images, and shadow masks, which have 1330 training triplets and 540 testing triplets. Moreover, due to the color mismatch in ISTD, the ISTD+ has reduced the color inconsistency using an image augmentation method. On the other hand, the SRD dataset consists of 2680 training and 408 testing pairs of shadow and shadow-free images without shadow masks. We additionally use the detection results of DHAN \cite{cun2020towards} for SRD shadow masks.

\subsubsection{Implementation Details.}
Our proposed method is implemented in PyTorch with a single GPU (NVIDIA GeForce GTX 3090). In the experiments, we employ the Adam optimizer to train our network for over 200 epochs with a batch size of 8 and the input patch size of 256 × 256. The initial learning rate is set to 1e-3 and gradually decreases with a dynamic decay strategy. As for the data augmentation, we randomly adopt the cropping, rotating, and flipping operations during the training, to circumvent the overfitting problem.

\subsubsection{Evaluation Metrics.}
Following the previous works, we use the root mean square error (RMSE) in the LAB color space between the shadow-removed result and its ground truth to evaluate the performance. Note that the RMSE value is actually calculated by the mean absolute error (MAE) in this task. The values of RMSE are calculated at each pixel of the shadow region, non-shadow region, as well as the whole image, and a lower value indicates better performance. Furthermore, to verify the effectiveness of our algorithm more comprehensively, we additionally assess the experimental results with the peak signal-to-noise ratio (PSNR) and the structural similarity (SSIM). The higher, the better.

\subsection{Shadow Removal Evaluation on ISTD Dataset}
\setlength{\tabcolsep}{4pt}
\begin{table}[!b]
\begin{center}
\setlength{\abovecaptionskip}{-0.02cm}
\setlength{\belowcaptionskip}{-0.01cm}
\caption{Quantitative shadow removal results of our network compared to state-of-the-art shadow removal methods on the ISTD dataset. The outcomes of the methods marked with an asterisk “$ * $" are referenced from their original papers.}
\label{table:ISTD_results}
\begin{tabular}{c|ccc|ccc|ccc}
\hline
\noalign{\smallskip}
\multirow{2}{*}{Method} & \multicolumn{3}{c}{RMSE} & \multicolumn{3}{c}{PSNR (dB)} & \multicolumn{3}{c}{SSIM} \\
~ & S & NS & ALL & S & NS & ALL & S & NS & ALL\\
\noalign{\smallskip}
\hline
\noalign{\smallskip}
Input Image & 32.11 & 6.83 & 10.97 & 22.40 & 27.30 & 20.56 & 0.936 & 0.975 & 0.892 \\
\noalign{\smallskip}
\hline
\noalign{\smallskip}
ST-CGAN \cite{wang2018stacked} & 9.55 & 6.13 & 6.69 & 33.73 & 29.50 & 27.43 & 0.981 & 0.957 & 0.928 \\
DSC \cite{hu2019direction} & 8.50 & \underline{5.13} & \underline{5.68} & 34.64 & \underline{31.22} & 28.97 & 0.983 & 0.968 & 0.943 \\
DHAN \cite{cun2020towards} & \underline{7.55} & 5.39 & 5.74 & \underline{35.52} & 31.01 & \underline{29.08} & \underline{0.988} & 0.969 & \underline{0.953} \\
RIS-GAN$ * $ \cite{zhang2020ris} & 8.99 & 6.33 & 6.95 & \multicolumn{6}{c}{---}\\
DC-SNet \cite{jin2021dc} & 10.57 & 5.83 & 6.60 & 31.68 & 28.98 & 26.37 & 0.976 & 0.957 & 0.921 \\
CANet$ * $ \cite{chen2021canet} & 8.86 & 6.07 & 6.15 & \multicolumn{6}{c}{---}\\
Fu \emph{et al.} \cite{fu2021auto} & 7.77 & 5.57 & 5.93 & 34.71 & 28.60 & 27.19 & 0.975 & 0.880 & 0.845 \\
BMNet \cite{zhu2022bijective} & 7.89 & 5.30 & 5.73 & 34.58 & 30.85 & 28.70 & \underline{0.988} & \underline{0.970} & 0.950 \\
\noalign{\smallskip}
\hline
\noalign{\smallskip}
CNSNet (Ours) & \textbf{6.56} & \textbf{4.23} & \textbf{4.61} & \textbf{36.67} & \textbf{32.15} & \textbf{30.29} & \textbf{0.991} & \textbf{0.979} & \textbf{0.965} \\
\noalign{\smallskip}
\hline
\end{tabular}
\end{center}
\end{table}
\setlength{\tabcolsep}{1.4pt}

As shown in Tables~\ref{table:ISTD_results} and \ref{table:ISTD+_results}, we report the quantitative shadow removal results of our CNSNet on the ISTD and ISTD+ datasets, and compare it with recent state-of-the-art (SOTA) algorithms, including ST-CGAN \cite{wang2018stacked}, DSC \cite{hu2019direction}, DHAN \cite{cun2020towards}, RIS-GAN \cite{zhang2020ris}, DC-ShadowNet \cite{jin2021dc}, G2R \cite{liu2021shadow}, CANet \cite{chen2021canet}, Fu \emph{et al.} \cite{fu2021auto}, and BMNet \cite{zhu2022bijective}. In addition, we compare the network parameters (Param.) and floating point operations (FLOPs) in Table~\ref{table:ISTD+_results}, where the values of Param. and FLOPs are directly referred from \cite{zhu2022bijective}. For the sake of fairness of comparison, these statistics are calculated from the de-shadowing results with a resolution of 256 $\times$ 256, presented by the authors or directly acquired from the original papers. In the following tables, S, NS, and ALL indicate the shadow region, non-shadow region, and entire image, respectively. Note that RMSE is calculated by averaging the RMSE over all the pixels in certain regions of the whole testing set, not per image. The first row (input image) shows the metrics of the original corresponding pair images of shadow and shadow-free, as a blank-control group. The best and the second-place values for each metric are respectively highlighted in \textbf{bold} and \underline{underlined}.

\begin{figure}[!t]
\centering
\includegraphics[height=5.5cm]{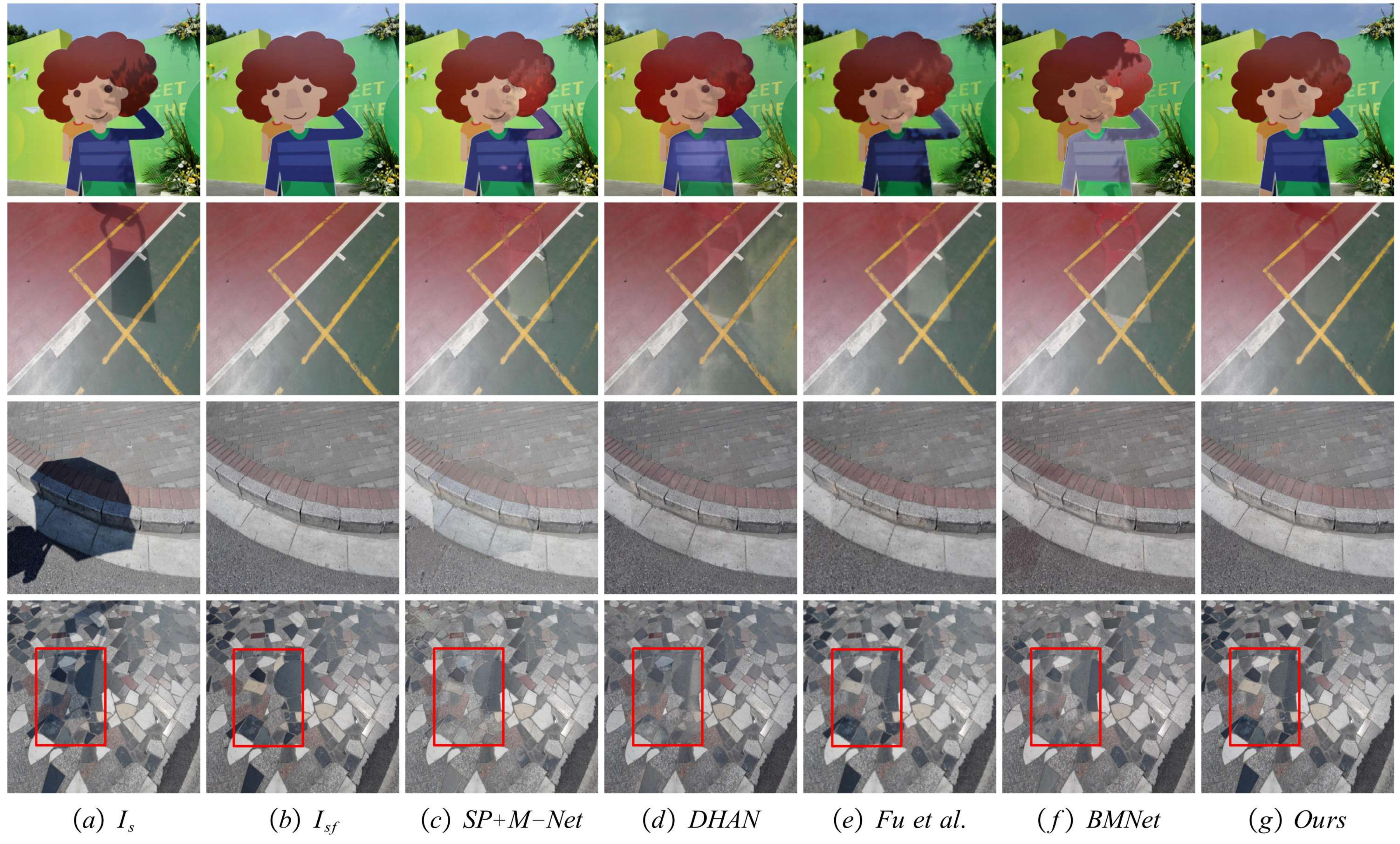}
\setlength{\abovecaptionskip}{-0.01cm}
\setlength{\belowcaptionskip}{-0.2cm}
\caption{Visual comparison results of shadow removal on the ISTD dataset. (a) Input shadow images, (b) corresponding ground-truth shadow-free images, and (c)-(g) results from SP+M-Net \cite{le2019shadow,le2021physics}, DHAN \cite{cun2020towards}, Fu \emph{et al.} \cite{fu2021auto}, BMNet \cite{zhu2022bijective}, and our CNSNet.}
\label{fig:ISTD_results}
\end{figure}

\setlength{\tabcolsep}{4pt}
\begin{table}[!b]
\begin{center}
\setlength{\abovecaptionskip}{-0.01cm}
\caption{Quantitative shadow removal results of our network compared to state-of-the-art shadow removal methods on the ISTD+ dataset.}
\label{table:ISTD+_results}
\begin{tabular}{c|ccc|ccc|c|c}
\hline
\noalign{\smallskip}
\multirow{2}{*}{Method} & \multicolumn{3}{c}{RMSE} & \multicolumn{3}{c}{PSNR (dB)} & Param. & FLOPs \\
~ & S & NS & ALL & S & NS & ALL & (M:$10^{6}$) & (G:$10^{9}$) \\
\noalign{\smallskip}
\hline
\noalign{\smallskip}
Input Image & 39.04 & 2.52 & 8.50 & 20.83 & 37.34 & 20.45 & \multicolumn{2}{c}{---}\\
\noalign{\smallskip}
\hline
\noalign{\smallskip}
DSC \cite{hu2019direction} & 7.54 & 3.16 & 3.88 & 35.97 & 35.76 & 32.05 & 22.30 & 123.47 \\
DHAN \cite{cun2020towards} & 11.30 & 7.17 & 7.85 & 32.91 & 27.14 & 25.65 & 21.75 & 262.87 \\
DC-SNet \cite{jin2021dc} & 10.43 & 3.68 & 4.78 & 32.00 & 33.53 & 28.76 & 21.16 & 105.00 \\
G2R \cite{liu2021shadow} & 7.41 & 3.03 & 3.74 & 35.76 & 35.54 & 31.88 & 22.76 & 113.87 \\
Fu \emph{et al.} \cite{fu2021auto} & 6.58 & 3.83 & 4.28 & 36.04 & 31.15 & 29.44 & 143.01 & 160.32 \\
BMNet \cite{zhu2022bijective} & \underline{5.70} & \underline{2.58} & \underline{3.09} & \underline{37.85} & \underline{37.39} & \underline{33.91} & \textbf{0.37} & \textbf{10.99} \\
\noalign{\smallskip}
\hline
\noalign{\smallskip}
CNSNet (Ours) & \textbf{5.60} & \textbf{2.47} & \textbf{2.98} & \textbf{38.10} & \textbf{37.74} & \textbf{34.20} & \underline{1.17} & \underline{17.67} \\
\noalign{\smallskip}
\hline
\end{tabular}
\end{center}
\vspace{-0.6cm}
\end{table}
\setlength{\tabcolsep}{1.4pt}

From Tables~\ref{table:ISTD_results} and \ref{table:ISTD+_results}, we can observe that our CNSNet achieves the best shadow removal performance than other SOTA methods by a large margin. The visualization results of the shadow removal comparison are displayed in Fig.~\ref{fig:ISTD_results}, which further justifies the effectiveness of our method. We produce a better visual restoration effect with fewer artifacts and boundary traces between the shadow and non-shadow regions. Specifically, although BMNet \cite{zhu2022bijective} has slightly fewer network parameters and FLOPs, the values of RMSE, PSNR, SSIM in our algorithm are significantly improved by 1.12, 1.59 dB, 0.015 in entire images, 1.33, 2.09 dB, 0.003 in shadow regions and 1.07, 1.30 dB, 0.009 in non-shadow regions. Besides, on the ISTD+ dataset, our method also has a great improvement, outperforming the BMNet \cite{zhu2022bijective} by 0.10, 0.11, 0.11 lower RMSE values and 0.25 dB, 0.35 dB, 0.29 dB higher PSNR values in shadow regions, non-shadow regions, and whole images, respectively. From the overall view, we effectively ensure the good balance of network parameters, efficiency, and performance.

\subsection{Shadow Removal Evaluation on SRD Dataset}
In Table~\ref{table:SRD_results}, we further report the shadow removal results on the SRD dataset. The compared baseline methods include DSC \cite{hu2019direction}, DHAN \cite{cun2020towards}, RIS-GAN \cite{zhang2020ris}, DC-ShadowNet \cite{jin2021dc}, CANet \cite{chen2021canet}, Fu \emph{et al.} \cite{fu2021auto}, and BMNet \cite{zhu2022bijective}. Our CNSNet still presents a competitive de-shadowing performance by decreasing the total RMSE value from 4.46 to 4.29, and increasing the PSNR value of the shadow regions from 35.05 dB to 35.10 dB.

\setlength{\tabcolsep}{4pt}
\begin{table}[!h]
\begin{center}
\setlength{\abovecaptionskip}{-0.02cm}
\caption{Quantitative shadow removal results of our network compared to state-of-the-art shadow removal methods on the SRD dataset.}
\vspace{-0.3cm}
\label{table:SRD_results}
\begin{tabular}{c|ccc|ccc|ccc}
\hline
\noalign{\smallskip}
\multirow{2}{*}{Method} & \multicolumn{3}{c}{RMSE} & \multicolumn{3}{c}{PSNR (dB)} & \multicolumn{3}{c}{SSIM} \\
~ & S & NS & ALL & S & NS & ALL & S & NS & ALL\\
\noalign{\smallskip}
\hline
\noalign{\smallskip}
Input Image & 39.31 & 4.54 & 14.11 & 18.96 & 31.44 & 18.19 & 0.871 & 0.975 & 0.829 \\
\noalign{\smallskip}
\hline
\noalign{\smallskip}
DSC \cite{hu2019direction} & 9.31 & \underline{3.46} & 5.07 & 32.20 & 34.90 & 29.87 & 0.969 & \underline{0.984} & 0.943 \\
DHAN \cite{cun2020towards} & 7.77 & 3.49 & 4.67 & 33.83 & 35.02 & 30.72 & 0.980 & \underline{0.984} & \underline{0.957} \\
RIS-GAN$ * $ \cite{zhang2020ris} & 8.22 & 6.05 & 6.78 & \multicolumn{6}{c}{---}\\
DC-SNet \cite{jin2021dc} & 8.28 & 3.71 & 4.97 & 33.40 & 34.93 & 30.55 & 0.974 & 0.983 & 0.947 \\
CANet$ * $ \cite{chen2021canet} & 7.82 & 5.88 & 5.98 & \multicolumn{6}{c}{---}\\
Fu \emph{et al.} \cite{fu2021auto} & 8.93 & 5.26 & 6.27 & 32.43 & 30.83 & 27.96 & 0.968 & 0.950 & 0.901 \\
BMNet$ * $ \cite{zhu2022bijective} & \textbf{6.61} & 3.61 & \underline{4.46} & \underline{35.05} & \textbf{36.02} & \textbf{31.69} & \underline{0.981} & 0.982 & 0.956 \\
\noalign{\smallskip}
\hline
\noalign{\smallskip}
CNSNet (Ours) & \underline{6.92} & \textbf{3.29} & \textbf{4.29} & \textbf{35.10} & \underline{35.69} & \textbf{31.69} & \textbf{0.982} & \textbf{0.986} & \textbf{0.959} \\
\noalign{\smallskip}
\hline
\end{tabular}
\end{center}
\vspace{-0.9cm}
\end{table}
\setlength{\tabcolsep}{1.4pt}

\subsection{Ablation Studies}
\label{section:Ablation}
In this subsection, we conduct several ablation studies on the ISTD dataset to demonstrate the contribution of each essential component in our framework. The notations used are listed as follows:
\begin{itemize}
\item “Ours (default)”: Taking the final results as the default control group;
\item “Ours w/o $ \mathcal{L} $”: Removing the certain loss term;
\item “Ours w/o SOAN/SAAT”: Removing the SOAN/SAAT module;
\item “$ SOAN_{BN} $”: Utilizing BN in SOAN instead of regional IN;
\item “$ SOAN_{IN} $”: Utilizing direct IN in SOAN instead of regional IN; and 
\item “$ SAAT_{hardmask} $”: Utilizing the guidance of hard masks in SAAT instead of soft masks.
\end{itemize}

\setlength{\tabcolsep}{4pt}
\begin{table}[!t]
\begin{center}
\caption{Ablation studies on choosing the loss functions and variants of the two key modules in our proposed CNSNet on the ISTD dataset.}
\label{table:Ablation_module}
\begin{tabular}{c|ccc|ccc}
\hline
\noalign{\smallskip}
\multirow{2}{*}{Method} & \multicolumn{3}{c}{RMSE} & \multicolumn{3}{c}{PSNR (dB)} \\
~ & S & NS & ALL & S & NS & ALL \\
\noalign{\smallskip}
\hline
\noalign{\smallskip}
Ours (default) & \textbf{6.56} & 4.23 & 4.61 & \textbf{36.67} & 32.15 & 30.29 \\
\noalign{\smallskip}
\hline
\noalign{\smallskip}
Ours w/o $ \mathcal{L}_{soft} $ & 7.54 & 4.66 & 5.13 & 35.38 & 31.51 & 29.48 \\
Ours w/o $ \mathcal{L}_{grad} $ & 6.91 & 3.97 & 4.45 & 36.56 & 33.15 & 30.91 \\
Ours w/o $ \mathcal{L}_{per} $ & 7.05 & 4.26 & 4.71 & 36.35 & 32.54 & 30.27 \\
\noalign{\smallskip}
\hline
\noalign{\smallskip}
Ours w/o SOAN & 7.54 & 4.43 & 4.94 & 35.39 & 31.80 & 29.75 \\
$ SOAN_{BN} $ & 7.23 & 4.24 & 4.73 & 36.43 & 32.30 & 30.18 \\
$ SOAN_{IN} $ & 7.17 & 4.07 & 4.58 & 35.85 & 33.12 & 30.53 \\
\noalign{\smallskip}
\hline
\noalign{\smallskip}
Ours w/o SAAT & 7.49 & 5.04 & 5.44 & 36.09 & 30.08 & 28.70 \\
$ SAAT_{HardMask} $ & 6.95 & 4.01 & 4.49 & 36.32 & 33.31 & 30.91 \\
\noalign{\smallskip}
\hline
\end{tabular}
\end{center}
\vspace{-0.6cm}
\end{table}
\setlength{\tabcolsep}{1.4pt}

In Table~\ref{table:Ablation_module}, we first justify the effects of the loss functions. It can be seen that $ \mathcal{L}_{soft} $ enables the network to acquire more accurate soft-region masks for better pixel-to-pixel connection, while $ \mathcal{L}_{grad} $ helps to balance the difference between the shadow and non-shadow regions for smooth recovery on the shadow boundaries, due to the dilated masks. Then, we investigate the performance gain brought by our SOAN module compared to other normalization methods (i.e., BN \cite{ioffe2015batch} and IN \cite{chen2021hinet,ulyanov2016instance}). Obviously, applying the SOAN block can maintain the statistical consistency of deep features between the two regions, thereby improving the de-shadowing quality of shadow regions by a large margin. Finally, in the SAAT module, we verify the superior performance of the generated soft-region masks over the hard masks, solving the problem of lost information transmission caused by absolute regional separation. 

\section{Conclusions}
In this paper, we develop a cleanness-navigated-shadow network (CNSNet) to achieve shadow removal via the short-range and long-range modeling. Our CNSNet exploits the auxiliary guidance of shadow masks to thoroughly investigate the regional information in both region-wise and pixel-wise ways through two novel modules, i.e., shadow-oriented adaptive normalization (SOAN) and shadow-aware aggregation with transformer (SAAT). The SOAN module keeps the statistical consistency by applying the information from the shadow-free region to the shadow region, while the SAAT module builds up the pixel-to-pixel connection between the two regions. Comprehensive experimental results have demonstrated the efficacy and superiority of our method, maintaining the balance of network complexity and performance at the meanwhile.
\\

\noindent\textbf{Acknowledgments}
This work was supported by the Anhui Provincial Natural Science Foundation under Grant 2108085UD12. We acknowledge the support of GPU cluster built by MCC Lab of Information Science and Technology Institution, USTC.
\clearpage
%
%
\bibliographystyle{splncs04}
\bibliography{mybibfile.bib}

\begin{thebibliography}{10}
\providecommand{\url}[1]{\texttt{#1}}
\providecommand{\urlprefix}{URL }
\providecommand{\doi}[1]{https://doi.org/#1}

\bibitem{chen2021pre}
Chen, H., Wang, Y., Guo, T., Xu, C., Deng, Y., Liu, Z., Ma, S., Xu, C., Xu, C.,
  Gao, W.: Pre-trained image processing transformer. In: Proceedings of the
  IEEE/CVF Conference on Computer Vision and Pattern Recognition. pp.
  12299--12310 (2021)

\bibitem{chen2021hinet}
Chen, L., Lu, X., Zhang, J., Chu, X., Chen, C.: Hinet: Half instance
  normalization network for image restoration. In: Proceedings of the IEEE/CVF
  Conference on Computer Vision and Pattern Recognition. pp. 182--192 (2021)

\bibitem{chen2021canet}
Chen, Z., Long, C., Zhang, L., Xiao, C.: Canet: A context-aware network for
  shadow removal. In: Proceedings of the IEEE/CVF International Conference on
  Computer Vision. pp. 4743--4752 (2021)

\bibitem{cucchiara2003detecting}
Cucchiara, R., Grana, C., Piccardi, M., Prati, A.: Detecting moving objects,
  ghosts, and shadows in video streams. IEEE Transactions on Pattern Analysis
  and Machine Intelligence  \textbf{25}(10),  1337--1342 (2003)

\bibitem{cun2020towards}
Cun, X., Pun, C.M., Shi, C.: Towards ghost-free shadow removal via dual
  hierarchical aggregation network and shadow matting gan. In: Proceedings of
  the AAAI Conference on Artificial Intelligence. vol.~34, pp. 10680--10687
  (2020)

\bibitem{ding2019argan}
Ding, B., Long, C., Zhang, L., Xiao, C.: Argan: Attentive recurrent generative
  adversarial network for shadow detection and removal. In: Proceedings of the
  IEEE/CVF International Conference on Computer Vision. pp. 10213--10222 (2019)

\bibitem{dosovitskiy2020image}
Dosovitskiy, A., Beyer, L., Kolesnikov, A., Weissenborn, D., Zhai, X.,
  Unterthiner, T., Dehghani, M., Minderer, M., Heigold, G., Gelly, S., et~al.:
  An image is worth 16x16 words: Transformers for image recognition at scale.
  arXiv preprint arXiv:2010.11929  (2020)

\bibitem{finlayson2009entropy}
Finlayson, G.D., Drew, M.S., Lu, C.: Entropy minimization for shadow removal.
  International Journal of Computer Vision  \textbf{85}(1),  35--57 (2009)

\bibitem{finlayson2005removal}
Finlayson, G.D., Hordley, S.D., Lu, C., Drew, M.S.: On the removal of shadows
  from images. IEEE Transactions on Pattern Analysis and Machine Intelligence
  \textbf{28}(1),  59--68 (2005)

\bibitem{fu2021auto}
Fu, L., Zhou, C., Guo, Q., Juefei-Xu, F., Yu, H., Feng, W., Liu, Y., Wang, S.:
  Auto-exposure fusion for single-image shadow removal. In: Proceedings of the
  IEEE/CVF Conference on Computer Vision and Pattern Recognition. pp.
  10571--10580 (2021)

\bibitem{gao2022towards}
Gao, J., Zheng, Q., Guo, Y.: Towards real-world shadow removal with a shadow
  simulation method and a two-stage framework. In: Proceedings of the IEEE/CVF
  Conference on Computer Vision and Pattern Recognition. pp. 599--608 (2022)

\bibitem{gong2016interactive}
Gong, H., Cosker, D.: Interactive removal and ground truth for difficult shadow
  scenes. Journal of the Optical Society of America A  \textbf{33}(9),
  1798--1811 (2016)

\bibitem{gryka2015learning}
Gryka, M., Terry, M., Brostow, G.J.: Learning to remove soft shadows. ACM
  Transactions on Graphics  \textbf{34}(5),  1--15 (2015)

\bibitem{guo2011single}
Guo, R., Dai, Q., Hoiem, D.: Single-image shadow detection and removal using
  paired regions. In: Proceedings of the IEEE Conference on Computer Vision and
  Pattern Recognition. pp. 2033--2040 (2011)

\bibitem{guo2012paired}
Guo, R., Dai, Q., Hoiem, D.: Paired regions for shadow detection and removal.
  IEEE Transactions on Pattern Analysis and Machine Intelligence
  \textbf{35}(12),  2956--2967 (2012)

\bibitem{hu2019direction}
Hu, X., Fu, C.W., Zhu, L., Qin, J., Heng, P.A.: Direction-aware spatial context
  features for shadow detection and removal. IEEE Transactions on Pattern
  Analysis and Machine Intelligence  \textbf{42}(11),  2795--2808 (2019)

\bibitem{hu2019mask}
Hu, X., Jiang, Y., Fu, C.W., Heng, P.A.: {Mask-ShadowGAN}: Learning to remove
  shadows from unpaired data. In: Proceedings of the IEEE/CVF International
  Conference on Computer Vision. pp. 2472--2481 (2019)

\bibitem{ioffe2015batch}
Ioffe, S., Szegedy, C.: Batch normalization: Accelerating deep network training
  by reducing internal covariate shift. In: Proceedings of the International
  Conference on Machine Learning. pp. 448--456 (2015)

\bibitem{jin2021dc}
Jin, Y., Sharma, A., Tan, R.T.: {DC-ShadowNet}: Single-image hard and soft
  shadow removal using unsupervised domain-classifier guided network. In:
  Proceedings of the IEEE/CVF International Conference on Computer Vision. pp.
  5027--5036 (2021)

\bibitem{jung2009efficient}
Jung, C.R.: Efficient background subtraction and shadow removal for
  monochromatic video sequences. IEEE Transactions on Multimedia
  \textbf{11}(3),  571--577 (2009)

\bibitem{le2019shadow}
Le, H., Samaras, D.: Shadow removal via shadow image decomposition. In:
  Proceedings of the IEEE/CVF International Conference on Computer Vision. pp.
  8578--8587 (2019)

\bibitem{le2020shadow}
Le, H., Samaras, D.: From shadow segmentation to shadow removal. In:
  Proceedings of the European Conference on Computer Vision. pp. 264--281
  (2020)

\bibitem{le2021physics}
Le, H., Samaras, D.: Physics-based shadow image decomposition for shadow
  removal. IEEE Transactions on Pattern Analysis and Machine Intelligence (01),
  ~1--1 (2021)

\bibitem{ling2021region}
Ling, J., Xue, H., Song, L., Xie, R., Gu, X.: Region-aware adaptive instance
  normalization for image harmonization. In: Proceedings of the IEEE/CVF
  Conference on Computer Vision and Pattern Recognition. pp. 9361--9370 (2021)

\bibitem{liu2021swin}
Liu, Z., Lin, Y., Cao, Y., Hu, H., Wei, Y., Zhang, Z., Lin, S., Guo, B.: Swin
  transformer: Hierarchical vision transformer using shifted windows. In:
  Proceedings of the IEEE/CVF International Conference on Computer Vision. pp.
  10012--10022 (2021)

\bibitem{liu2021shadow}
Liu, Z., Yin, H., Wu, X., Wu, Z., Mi, Y., Wang, S.: From shadow generation to
  shadow removal. In: Proceedings of the IEEE/CVF Conference on Computer Vision
  and Pattern Recognition. pp. 4927--4936 (2021)

\bibitem{nadimi2004physical}
Nadimi, S., Bhanu, B.: Physical models for moving shadow and object detection
  in video. IEEE Transactions on Pattern Analysis and Machine Intelligence
  \textbf{26}(8),  1079--1087 (2004)

\bibitem{perez2003poisson}
P{\'e}rez, P., Gangnet, M., Blake, A.: Poisson image editing. In: ACM SIGGRAPH,
  pp. 313--318 (2003)

\bibitem{qu2017deshadownet}
Qu, L., Tian, J., He, S., Tang, Y., Lau, R.W.: {DeshadowNet}: A multi-context
  embedding deep network for shadow removal. In: Proceedings of the IEEE
  Conference on Computer Vision and Pattern Recognition. pp. 4067--4075 (2017)

\bibitem{ronneberger2015u}
Ronneberger, O., Fischer, P., Brox, T.: {U-Net}: Convolutional networks for
  biomedical image segmentation. In: Proceedings of the International
  Conference on Medical Image Computing and Computer-Assisted Intervention. pp.
  234--241 (2015)

\bibitem{sanin2010improved}
Sanin, A., Sanderson, C., Lovell, B.C.: Improved shadow removal for robust
  person tracking in surveillance scenarios. In: Proceedings of the 20th
  International Conference on Pattern Recognition. pp. 141--144 (2010)

\bibitem{shor2008shadow}
Shor, Y., Lischinski, D.: The shadow meets the mask: Pyramid-based shadow
  removal. In: Computer Graphics Forum. vol.~27, pp. 577--586 (2008)

\bibitem{song2022vision}
Song, Y., He, Z., Qian, H., Du, X.: Vision transformers for single image
  dehazing. arXiv preprint arXiv:2204.03883  (2022)

\bibitem{ulyanov2016instance}
Ulyanov, D., Vedaldi, A., Lempitsky, V.: Instance normalization: The missing
  ingredient for fast stylization. arXiv preprint arXiv:1607.08022  (2016)

\bibitem{vasluianu2021shadow}
Vasluianu, F.A., Romero, A., Van~Gool, L., Timofte, R.: Shadow removal with
  paired and unpaired learning. In: Proceedings of the IEEE/CVF Conference on
  Computer Vision and Pattern Recognition. pp. 826--835 (2021)

\bibitem{vaswani2017attention}
Vaswani, A., Shazeer, N., Parmar, N., Uszkoreit, J., Jones, L., Gomez, A.N.,
  Kaiser, {\L}., Polosukhin, I.: Attention is all you need. Advances in Neural
  Information Processing Systems  \textbf{30} (2017)

\bibitem{vicente2017leave}
Vicente, T.F.Y., Hoai, M., Samaras, D.: Leave-one-out kernel optimization for
  shadow detection and removal. IEEE Transactions on Pattern Analysis and
  Machine Intelligence  \textbf{40}(3),  682--695 (2017)

\bibitem{wan2022crformer}
Wan, J., Yin, H., Wu, Z., Wu, X., Liu, Z., Wang, S.: {CRFormer}: A cross-region
  transformer for shadow removal. arXiv preprint arXiv:2207.01600  (2022)

\bibitem{wang2018stacked}
Wang, J., Li, X., Yang, J.: Stacked conditional generative adversarial networks
  for jointly learning shadow detection and shadow removal. In: Proceedings of
  the IEEE Conference on Computer Vision and Pattern Recognition. pp.
  1788--1797 (2018)

\bibitem{wang2021dynamic}
Wang, N., Zhang, Y., Zhang, L.: Dynamic selection network for image inpainting.
  IEEE Transactions on Image Processing  \textbf{30},  1784--1798 (2021)

\bibitem{wang2018non}
Wang, X., Girshick, R., Gupta, A., He, K.: Non-local neural networks. In:
  Proceedings of the IEEE Conference on Computer Vision and Pattern
  Recognition. pp. 7794--7803 (2018)

\bibitem{wang2022uformer}
Wang, Z., Cun, X., Bao, J., Zhou, W., Liu, J., Li, H.: Uformer: A general
  u-shaped transformer for image restoration. In: Proceedings of the IEEE/CVF
  Conference on Computer Vision and Pattern Recognition. pp. 17683--17693
  (2022)

\bibitem{wen2008example}
Wen, C.L., Hsieh, C.H., Chen, B.Y., Ouhyoung, M.: Example-based multiple local
  color transfer by strokes. In: Computer Graphics Forum. vol.~27, pp.
  1765--1772 (2008)

\bibitem{xiao2013fast}
Xiao, C., She, R., Xiao, D., Ma, K.L.: Fast shadow removal using adaptive
  multi-scale illumination transfer. In: Computer Graphics Forum. vol.~32, pp.
  207--218 (2013)

\bibitem{xu2022snr}
Xu, X., Wang, R., Fu, C.W., Jia, J.: {SNR}-aware low-light image enhancement.
  In: Proceedings of the IEEE/CVF Conference on Computer Vision and Pattern
  Recognition. pp. 17714--17724 (2022)

\bibitem{yarlagadda2018reflectance}
Yarlagadda, S.K., Zhu, F.: A reflectance based method for shadow detection and
  removal. In: Proceedings of the IEEE Southwest Symposium on Image Analysis
  and Interpretation. pp. 9--12 (2018)

\bibitem{yu2020region}
Yu, T., Guo, Z., Jin, X., Wu, S., Chen, Z., Li, W., Zhang, Z., Liu, S.: Region
  normalization for image inpainting. In: Proceedings of the AAAI Conference on
  Artificial Intelligence. vol.~34, pp. 12733--12740 (2020)

\bibitem{zamir2022restormer}
Zamir, S.W., Arora, A., Khan, S., Hayat, M., Khan, F.S., Yang, M.H.: Restormer:
  Efficient transformer for high-resolution image restoration. In: Proceedings
  of the IEEE/CVF Conference on Computer Vision and Pattern Recognition. pp.
  5728--5739 (2022)

\bibitem{zhang2020ris}
Zhang, L., Long, C., Zhang, X., Xiao, C.: {RIS-GAN}: Explore residual and
  illumination with generative adversarial networks for shadow removal. In:
  Proceedings of the AAAI Conference on Artificial Intelligence. vol.~34, pp.
  12829--12836 (2020)

\bibitem{zhang2015shadow}
Zhang, L., Zhang, Q., Xiao, C.: Shadow remover: Image shadow removal based on
  illumination recovering optimization. IEEE Transactions on Image Processing
  \textbf{24}(11),  4623--4636 (2015)

\bibitem{zhang2018improving}
Zhang, W., Zhao, X., Morvan, J.M., Chen, L.: Improving shadow suppression for
  illumination robust face recognition. IEEE Transactions on Pattern Analysis
  and Machine Intelligence  \textbf{41}(3),  611--624 (2018)

\bibitem{zhu2020sean}
Zhu, P., Abdal, R., Qin, Y., Wonka, P.: Sean: Image synthesis with semantic
  region-adaptive normalization. In: Proceedings of the IEEE/CVF Conference on
  Computer Vision and Pattern Recognition. pp. 5104--5113 (2020)

\bibitem{zhu2022bijective}
Zhu, Y., Huang, J., Fu, X., Zhao, F., Sun, Q., Zha, Z.J.: Bijective mapping
  network for shadow removal. In: Proceedings of the IEEE/CVF Conference on
  Computer Vision and Pattern Recognition. pp. 5627--5636 (2022)

\end{thebibliography}
\end{document}